\documentclass[letterpaper, 10 pt, conference]{ieeeconf}  

\IEEEoverridecommandlockouts                              

\usepackage{amsmath,amsfonts}
\usepackage{algorithmic}
\usepackage{algorithm}
\usepackage{amsmath}
\usepackage{dirtytalk}
\usepackage{array}
\usepackage{textcomp}
\usepackage{url}
\usepackage{verbatim}
\usepackage{multirow}
\usepackage{graphicx}
\usepackage[caption=false,font=footnotesize]{subfig}
\usepackage{xcolor}
\usepackage{cite}
\usepackage{makecell}
\usepackage{tabularx}
\usepackage{booktabs}
\usepackage{siunitx}
\usepackage{import}
\usepackage{bm}
 
\usepackage{censor}
\usepackage{soul}
\usepackage{color}
\sethlcolor{black}

\title{\LARGE \bf
Exoskeleton Control through Learning to Reduce Biological Joint Moments in Simulations
}

\author{
        Zihang You and Xianlian Zhou
\thanks{Zihang You, and Xianlian Zhou are with the Department of Biomedical Engineering, New Jersey Institute of Technology, Newark, NJ 07102, USA (zy4@njit.edu, alexzhou@njit.edu)}
}

\begin{document}

\maketitle
\thispagestyle{empty}
\pagestyle{empty}

\begin{abstract}
Data-driven joint-moment predictors offer a scalable alternative to laboratory-based inverse-dynamics pipelines for biomechanics estimation and exoskeleton control. Meanwhile, physics-based reinforcement learning (RL) enables simulation-trained controllers to learn dynamics-aware assistance strategies without extensive human experimentation. However, quantitative verification of simulation-trained exoskeleton torque predictors, and their impact on human joint power injection, remains limited. This paper presents (1) an RL framework to learn exoskeleton assistance policies that reduce biological joint moments, and (2) a validation pipeline that verifies the trained control networks using an open-source gait dataset through inference and comparison with biological joint moments. Simulation-trained multilayer perceptron (MLP) controllers are developed for level-ground and ramp walking, mapping short-horizon histories of bilateral hip and knee kinematics to normalized assistance torques.
Results show that predicted assistance preserves task-intensity trends across speeds and inclines. Agreement is particularly strong at the hip, with cross-correlation coefficients reaching 0.94 at 1.8\,m/s and 0.98 during 5° decline walking, demonstrating near-matched temporal structure. Discrepancies increase at higher speeds and steeper inclines, especially at the knee, and are more pronounced in joint power comparisons.
Delay tuning biases assistance toward greater positive power injection; modest timing shifts increase positive power and improve agreement in specific gait intervals. 
Together, these results establish a quantitative validation framework for simulation-trained exoskeleton controllers, demonstrate strong sim-to-data consistency at the torque level, and highlight both the promise and the remaining challenges for sim-to-real transfer.

\end{abstract}

\section{INTRODUCTION}

In recent years, data-driven neural network (NN) models have been widely adopted to rapidly estimate lower limb joint dynamics from kinematic and wearable signals, enabling low-cost and scalable gait analysis and robotic control\cite{belal2024deep, lim2019prediction}. Meanwhile, reinforcement learning (RL) in physics-based simulation can learn dynamics-aware representations and control policies without extensive real-world experimentation\cite{su2023simulating, luo2021reinforcement}.
Lower limb exoskeletons typically provide stable or periodic assistance across different gait activities such as walking, running, and climbing stairs.  Due to the periodic nature of gait, joint moments are commonly regarded as the primary reference for constructing assistive torque strategies and closed-loop control laws\cite{baud2021review}. Recent studies have shown that when a system can obtain or estimate users’ instantaneous joint moments in real time, these signals can be directly incorporated into unified assistance control frameworks, enabling adaptive torque output across different gait conditions while reducing metabolic cost\cite{zhang2025joint, molinaro2024estimating}. Similarly, estimating biological hip joint moments using mechanical sensors and deep learning models has been explicitly proposed as a continuous reference signal for modulating exoskeleton assistance intensity, thereby improving usability in real-world environments\cite{molinaro2020biological, mccabe2023developing}. 
Beyond joint moments, joint power provides a more direct dynamical perspective on energy generation and absorption. Therefore, when evaluating dynamics prediction models or control strategies, jointly analyzing both joint moments and the delivered joint power not only captures trends in torque magnitude and timing but also reveals how these trends propagate at the energetic level and potentially influence assistive performance.

Although experimental musculoskeletal modeling combined with inverse dynamics provides mature and accurate estimates of human biomechanical states, this workflow relies heavily on specialized laboratory equipment and complex protocols\cite{zhang2024integrating, rose2022model}, which limits scalability. Therefore, researchers have begun leveraging more accessible mechanical sensor signals as inputs to neural networks to directly predict lower limb joint moments, thereby bypassing costly motion capture or force plate measurements and improving deployment feasibility. Using IMU as an example, prior studies have systematically evaluated multiple deep learning architectures, including convolutional, recurrent, and temporal Transformer models, for predicting hip, knee, and ankle joint moments individually from IMU acceleration and angular velocity, demonstrating the potential of this approach\cite{altai2023performance}. Further work has shown that even a single IMU combined with an attention-based CNN–BiLSTM can achieve high correlations in joint moment estimation across diverse walking conditions\cite{liang2023deep}. 
While these data-driven approaches offer a pathway for translating joint dynamics from laboratory settings to real-world applications, their performance remains highly dependent on training data scale and coverage, and cross-domain deployment requires more systematic evaluation to quantify the gap between simulation and reality\cite{altai2023performance}.

Physics-based RL in simulation has emerged as a complementary approach for learning dynamics representations and locomotion controllers, motivated by the limitations of inverse-dynamics pipelines and data-driven joint-kinetics regression models\cite{huang2025lower, zhang2024integrating}. Deep RL methods such as proximal policy optimization combined with imitation priors have been shown to produce stable forward walking control in simulations of both able-bodied individuals and prosthesis users\cite{de2021deep}. Subsequent studies have further demonstrated that RL can generate experimentally plausible walking speeds within full musculoskeletal dynamics models without relying on reference trajectories, thereby enabling the exploration of how neuromuscular deficits manifest as gait alterations in computational simulations\cite{su2023simulating}. 
More recently, studies have reported exoskeleton assistance strategies learned entirely in simulation without experimental data, capable of automatically adapting to multiple activities and reducing metabolic cost, underscoring the practical potential of this paradigm\cite{luo2024experiment}. Another framework, Exo-plore, combines neuromechanical simulation with deep reinforcement learning to optimize hip exoskeleton assistance in simulation while modeling human adaptation and extending to pathological gait scenarios\cite{leem2026exo}.

In practice, torque-predictive networks, including exoskeleton assistance torque prediction networks, learned in physics-based RL simulations are not rigorously evaluated, e.g., using independent open-source datasets, and their errors are not well characterized in terms of amplitude, timing, and phase dependence. Moreover, most studies report joint-moment accuracy alone, while joint power often provides a more direct energetic interpretation, and the associated moment-to-power error propagation is seldom assessed under a consistent protocol. These limitations restrict the reliability of simulation-trained predictors for biomechanics interpretation and assistive control.

In this work, we first present a new method to learn exoskeleton assistance policies by minimizing biological joint moments in a physics-based RL simulation environment. We then propose a unified evaluation pipeline that assesses the learned control networks with a public dataset by performing inference and comparing predicted assistance torques with ground truth biological joint moments. Prediction fidelity is assessed at both the joint-moment level and the derived joint-power level against ground truth under consistent alignment and evaluation metrics. Furthermore, we evaluate the performance of the level-ground walking controller under ramp walking conditions to examine its generalization capability. Finally, we  investigated how artificially introduced joint torque delays reshape the inferred joint power patterns, providing actionable insights for translating simulated assistance strategies into energetically consistent exoskeleton control across varying walking speeds and inclines.

\section{Method}

\subsection{Reinforcement Learning based Exoskeleton Control with Neural Networks}
Here we present a new method to learn exoskeleton control policies for walking within a physics-based musculoskeletal simulation environment. The human model consists of 23 rigid body segments actuated by 304 Hill-type muscle–tendon units~\cite{park2022generative}. Muscles span anatomically defined origin–insertion sites and generate joint torques through force transmission along their paths. The framework allows subject-specific scaling of segment dimensions and modification of muscle parameters to represent variations in anthropometry or muscle function~\cite{park2022generative}.
To facilitate sim-to-real transfer and reduce dependence on device-specific properties, exoskeleton assistance was modeled as idealized joint torques applied directly at the biological hip and knee joints. Because the device joints are assumed to be rigidly aligned with the human joints, exoskeleton kinematics were approximated by the corresponding human joint angles and angular velocities. This abstraction enables implementation using wearable sensors (e.g., IMUs) mounted on the user and supports portability across different exoskeleton platforms while reducing simulation complexity.

The overall framework follows a simulation-based learning paradigm composed of a Human Control Network (HCN), a Muscle Coordination Network (MCN), and a dedicated Exoskeleton Control Network (ECN)\cite{luo2024experiment}. The HCN and MCN are consistent with formulations presented in \cite{ lee2019scalable,luo2023robust}, while the ECN is trained separately to generate assistive torques from joint kinematics. The ECN is implemented as a control policy,
\begin{equation}
a_e = \pi_{\psi_e}(s_e)
\label{eq:policy}
\end{equation}
where $s_e$ consists of a short time history (four consecutive time steps) of bilateral hip and knee joint angles and angular velocities (i.e., 4 joints × 2 kinematic variables × 4 time steps = 32), together with an additional scalar delay parameter (default to 25\,ms) appended to the state. The network comprises three fully connected hidden layers (64 neurons each) with ReLU activations and a Tanh output layer that produces normalized torque commands (${\tau}_{exo} = a_e$) within [-1,\,1] for both hips and knees. 

The ECN parameters are optimized via supervised learning using
\begin{equation}
\begin{aligned}
\mathcal{L}(\psi)
&= \mathbb{E}\!\Big[
\left\| \boldsymbol{\tau}_{d} - \boldsymbol{\tau}_{exo} \right\|^{2}
+ w_{\text{reg}}
\left\| \boldsymbol{\tau}_{exo} \right\|^{2}]
\end{aligned}
\label{eq:loss}
\end{equation}
where $\boldsymbol{\tau}_d$ denotes the target biological joint moments, normalized by the maximum assistance torque of $50 Nm$ and clipped to [-1,\,1], and $\boldsymbol{\tau}_{exo}$ denotes the ECN predicted, normalized exoskeleton torques. The regularization term
($w_{\text{reg}} = 0.01$) penalizes excessive torque magnitude to promote comfort and efficiency. Note before normalization and clipping, $\boldsymbol{\tau}_d$ is the desired human biological joint moment, predicted from HCN with an additional PD control\cite{ lee2019scalable}, after exoskeleton assistance applied. 

The HCN is trained with reinforcement learning using the PPO algorithm \cite{schulman2017proximal} and the gaitnet rewards in~\cite{park2022generative}.
During training, the gait stride and cadence were randomized with a target walking speed range from 0.65\,m/s to 2.0\,m/s under normal distribution; and the human model's height and weight were randomized between [53.6\,kg, 1.55\,m] and [97.1\,kg, 1.94\,m].
Using this framework, ECNs were trained for a hip–knee exoskeleton for level-ground walking, 5° and 10° incline and decline walking. The ECN processes $s_e$ at 100 Hz and outputs normalized torque commands within [-1,\,1]. 

\begin{figure}[t]
  \centering
  \includegraphics[width=\linewidth]{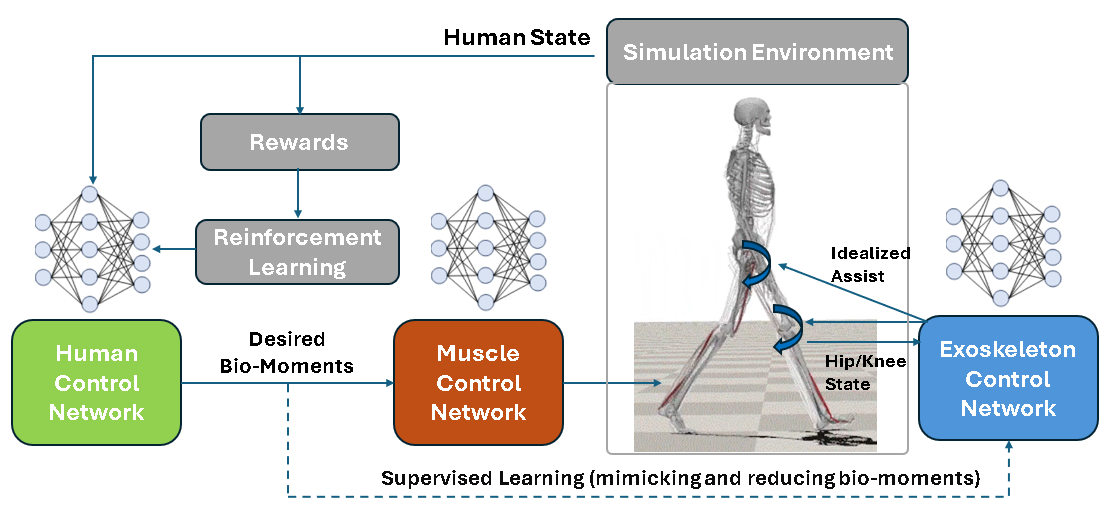}
  \caption{The RL learning framework with Exoskeleton Control Network (ECN) and Muscle Coordination Network (MCN). Idealized torque assistance (blue curved arrows) is used instead of modeling any specific physical devices. The Human Control Network (HCN) predicted action is transferred to the desired torques (i.e. human biological joint moments), which is affected by exoskeleton assistance and corresponds to the full total joint torque minus the assistance. The desired biological torques are fed to MCN to predict muscle activations. The ECN predicts normalized assistance torques and is trained in supervised learning to reduce the biological joint moments via the loss in Eq.~\ref{eq:loss}. The trained ECN and MCN work together to control the human model for walking on level ground and ramps.}
  \label{fig:rl_sim}
  \vspace{-0.2in}
\end{figure}

\subsection{Network Inference and Verification with an Open-Source Dataset}
In this section, we present an evaluation pipeline that verifies the trained exoskeleton control networks with an open-source gait dataset~\cite{molinaro2024task} through inference and comparison with biological joint moments. The dataset comprises three phases, and only the Phase 1 data, in which subjects wore a hip-knee exoskeleton without assistance enabled, were used for inference evaluation. The selected activities were consistent with those used during RL training, which are level-ground walking and ramp ascent and descent. Specifically, level-ground walking was performed at five different speeds (0.6\,m/s, 1.2\,m/s, 1.8\,m/s, 2.0\,m/s, and 2.5\,m/s), while ramp walking included two incline levels (5° and 10°). Each subject completed one walking trial for each speed and incline condition. The data from Subject 1 were excluded from inference evaluation due to data loss. In total, data from nine subjects were included in the inference tests. Within the dataset, exoskeleton encoders provide both hip and knee joint angles and angular velocities, while inverse kinematics (IK) provide only the angular data but velocities can be derived from differentiation. However, preliminary inspection revealed substantial discrepancies in angular velocity magnitude between these two sources. The angular velocities derived from IK exhibit ranges that are more consistent with physiological characteristics. Therefore, this study uses the IK data with joint angular velocities computed from joint angles using central finite differencing, for subsequent analysis. All raw data were originally sampled at 200 Hz and were resampled to 100 Hz to construct input required by the MLP model. 

We conducted inference evaluations under two scenarios. In the first scenario, joint angles and angular velocities for each activity were fed into the corresponding model to estimate exoskeleton assistance torques. The assistance power was then computed by multiplying the estimated assistance torque by the corresponding angular velocities provided in the open source dataset. In the second scenario to test the generalization capability of the level ground walking controller, joint angles and angular velocities from ramp walking were provided to the level-ground walking model to generate a mismatched inference torques, which were used to compute the corresponding mismatched joint power.

For statistical analysis, we computed the cross-correlation between the inferred assistance torque and the ground truth (GT) biological joint moments to quantify their level of similarity. Second, we calculated the mean power (MP), mean positive power(MPP), and mean negative power(MNP) for the hip and knee under each walking condition using the following formulation, in order to evaluate the energy implications of the inferred joint moments over the gait cycle. In (3) and (4), $n_p$ and $n_n$ denote the total number of positive and negative power data points, respectively. $n = n_p + n_n$ denotes the total number of data points. $\tau_i$ denotes the torque, and $\omega_i$ denotes the joint velocity.
\begin{align}
    &\text{MPP} = \frac{1}{n} \sum_{i=1}^{n_p} \tau_i \omega_i \quad (\text{for } \tau_i \omega_i > 0) \\
    &\text{MNP} = \frac{1}{n} \sum_{i=1}^{n_n} \tau_i \omega_i \quad (\text{for } \tau_i \omega_i < 0) \\
    &\text{MP} = \text{MPP} + \text{MNP}
\end{align}

Finally, we investigated the effect of introducing delay between the inferred torques and joint velocities on joint power estimation. In this evaluation, under 1.2\,m/s level-ground walking, 5° incline and decline walking conditions, we manually imposed delays of 50\,ms, 100\,ms, and 150\,ms on the inferred torque and compared the resulting power profiles with those obtained without delay and with the GT. These three walking conditions were selected because they more closely represent typical human walking speeds and commonly encountered ramp inclinations in real-world environments.

\section{Results}
\begin{figure}[t]
  \centering
  \includegraphics[width=0.9\linewidth]{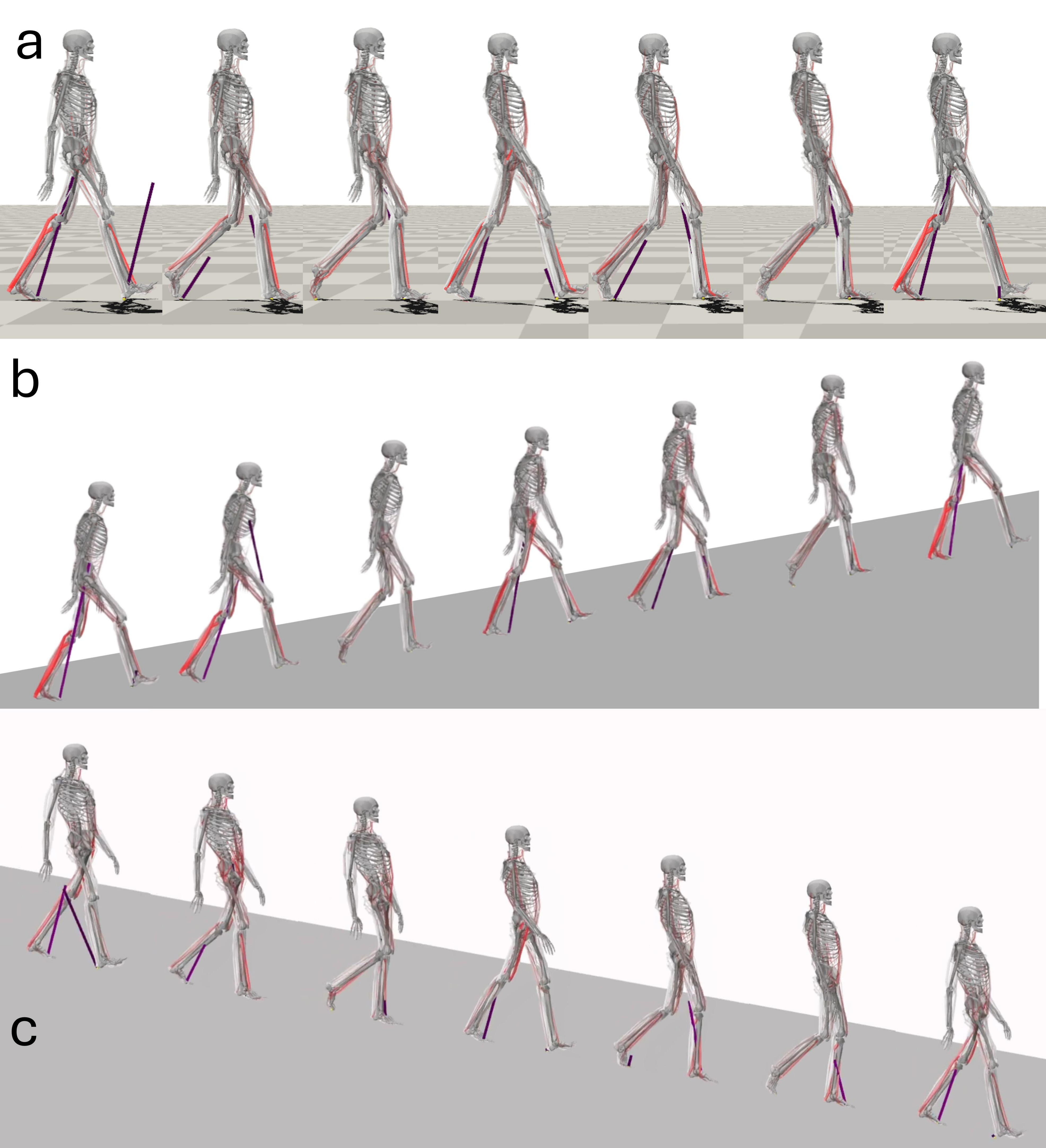}
  \caption{Simulation results of walking gait on (a)level ground, (b)inclined (10°) and (c)declined (10°) grade surfaces with idealized hip and knee assistance. The purple lines on the foot are ground reaction forces.}
  \label{fig:walking_sim}
  \vspace{-0.2in}
\end{figure}

Fig.~\ref{fig:walking_sim} presents snapshots of the gait events over one gait cycle with three human and exoskeleton controllers trained in the simulation environment applied to the musculoskeletal model, during level-ground walking, 10° incline walking, and 10° decline walking, respectively. In all scenarios, the musculoskeletal model exhibits smooth and natural gait patterns and the exoskeleton assistance reduced the human joint biological moments substantially in simulation.

The inferred assistance torque and power profiles were segmented into gait cycles and compared with the corresponding GT curves. For gait cycle segmentation, we used the maximum hip flexion angle as the starting point of gait phase as ground reaction forces are not available for some subjects and activities in the GT data. Typically, the maximum hip flexion occurs shortly before heel strike. Because the GT biological moments were expressed in Nm/kg, whereas the model output was normalized within the range of minus one to one, the inferred assistance torques were rescaled to match the amplitude range of the GT prior to visualization, preserving zero values.

\begin{figure}[htbp]
  \centering
  \includegraphics[width=1\linewidth]{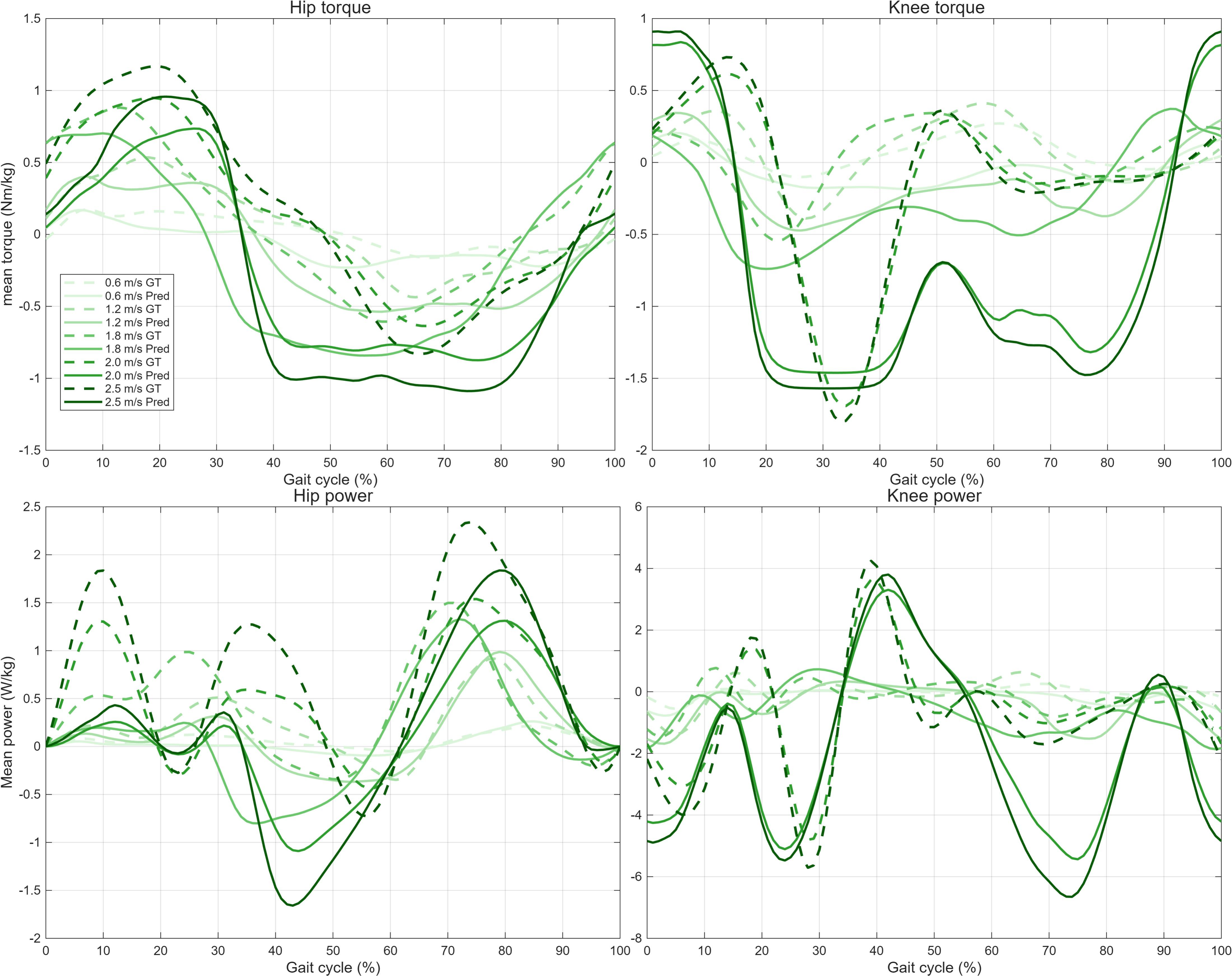}
  \caption{Inference hip and knee assistance torque and power (Pred) vs. ground truth (GT) biological moment and power under level-ground walking. The sign of hip and knee torques: hip flexion is negative and knee flexion is positive.}
  \label{fig:normal_walk_torque_power}
  \vspace{-0.2in}
\end{figure}

Fig.~\ref{fig:normal_walk_torque_power} illustrates the inferred hip and knee assistance torque and joint power curves in comparison with the corresponding GT curves across different level-ground walking speeds (0.6–2.5\,m/s). 
Hip assistance torque curves exhibited a clear speed dependence under both inference and GT conditions. The inferred torques preserved the steady increase of peak magnitude with speed, indicating consistent scaling with locomotion intensity. However, notable phase-specific discrepancies were observed. Specifically, the inferred profiles displayed a saturated negative region across 40–85\% (mid-stance to early swing) rather than the negative GT peak.

The inferred knee moments captured the general speed scaling and the timing of the pre-contact positive peak, but exhibited systematic phase distortions in stance. Specifically, the inferred profiles transitioned earlier into negative moments and formed a wider, flattened negative region spanning 15–40\% (loading response to mid-stance), whereas GT presented a more localized negative peak with a smoother recovery toward 80–95\% (mid- to terminal swing). 
These discrepancies were most evident at 2\,m/s and 2.5\,m/s, which were outside the range of velocity in simulation.


Across walking speeds, power from both GT and inference demonstrated strong speed scaling, with larger power excursions at higher walking speeds, confirming that the inferred torque and kinematics jointly reflect increasing energetic demand. For the hip, the computed power tended to under-represent the GT’s early positive contribution while producing more broader power excursions later in the cycle, yielding an overall shift toward wider generation and absorption episodes rather than sharper events happening in GT. For the knee, the mismatch was larger and more sensitive to speed, with higher-speed trials showing amplified positive and negative power variations relative to GT, consistent with an overestimation of both generation and absorption magnitudes when speed increased.

\begin{figure}[htbp]
  \centering
  \includegraphics[width=1\linewidth]{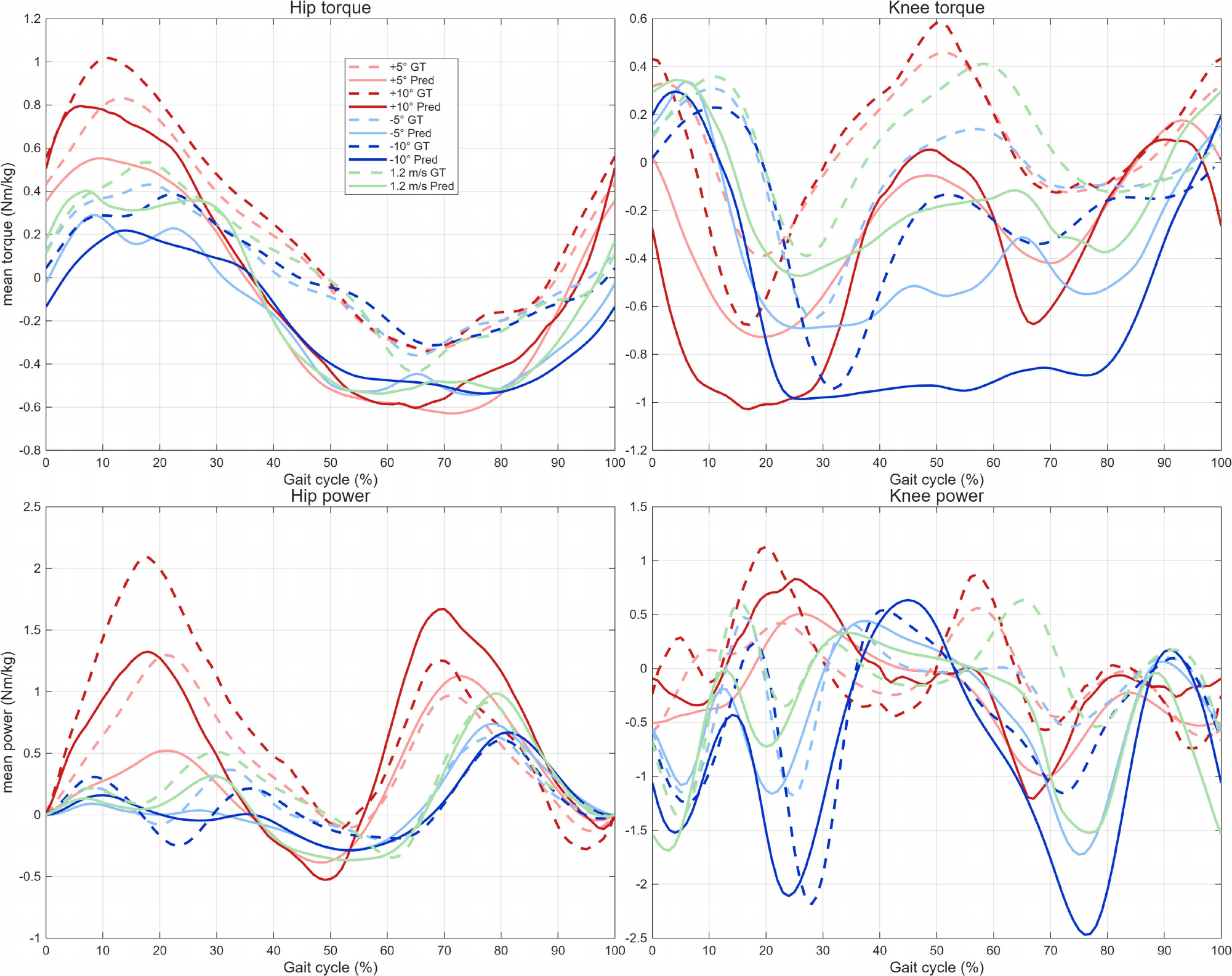}
  \caption{Inference hip and knee assistance torque and power vs. GT biological moment and power under ramp walking. Positive degree correspond to incline walking, negative degree denotes decline walking, and 1.2\,m/s represents the reference level-ground walking.}
  \label{fig:incline_walk_torque_power}
  \vspace{-0.2in}
\end{figure}

Fig.~\ref{fig:incline_walk_torque_power} illustrates the inferred hip and knee assistance torque and joint power profiles under different incline conditions, compared with the corresponding GT curves and the 1.2\,m/s level-ground walking condition. 
Overall, the inferred results preserved the general incline-dependent ordering, with incline walking characterized by greater hip extension demand and higher net positive power generation, while decline walking exhibited stronger knee flexion-dominant moments and more negative power absorption. For hip assistance torques, the models tended to underestimate the peak at extension and overestimate the peak at flexion, but the overall waveform trend remained consistent with GT, and sign transitions within the gait cycle were kept. For knee assistance torques, the results clearly distinguished between incline and decline walking. However, during decline walking, particularly at -10°, the model exhibited exaggerated and prolonged negative torque behavior compared with GT. In terms of hip joint power, the early-cycle positive power intensity was attenuated relative to GT. In contrast, knee joint power was most affected during decline walking, where negative absorption events were amplified and temporally broadened compared with GT, again most prominently under the -10° condition.

\begin{figure}[htbp]
  \centering
  \includegraphics[width=1\linewidth]{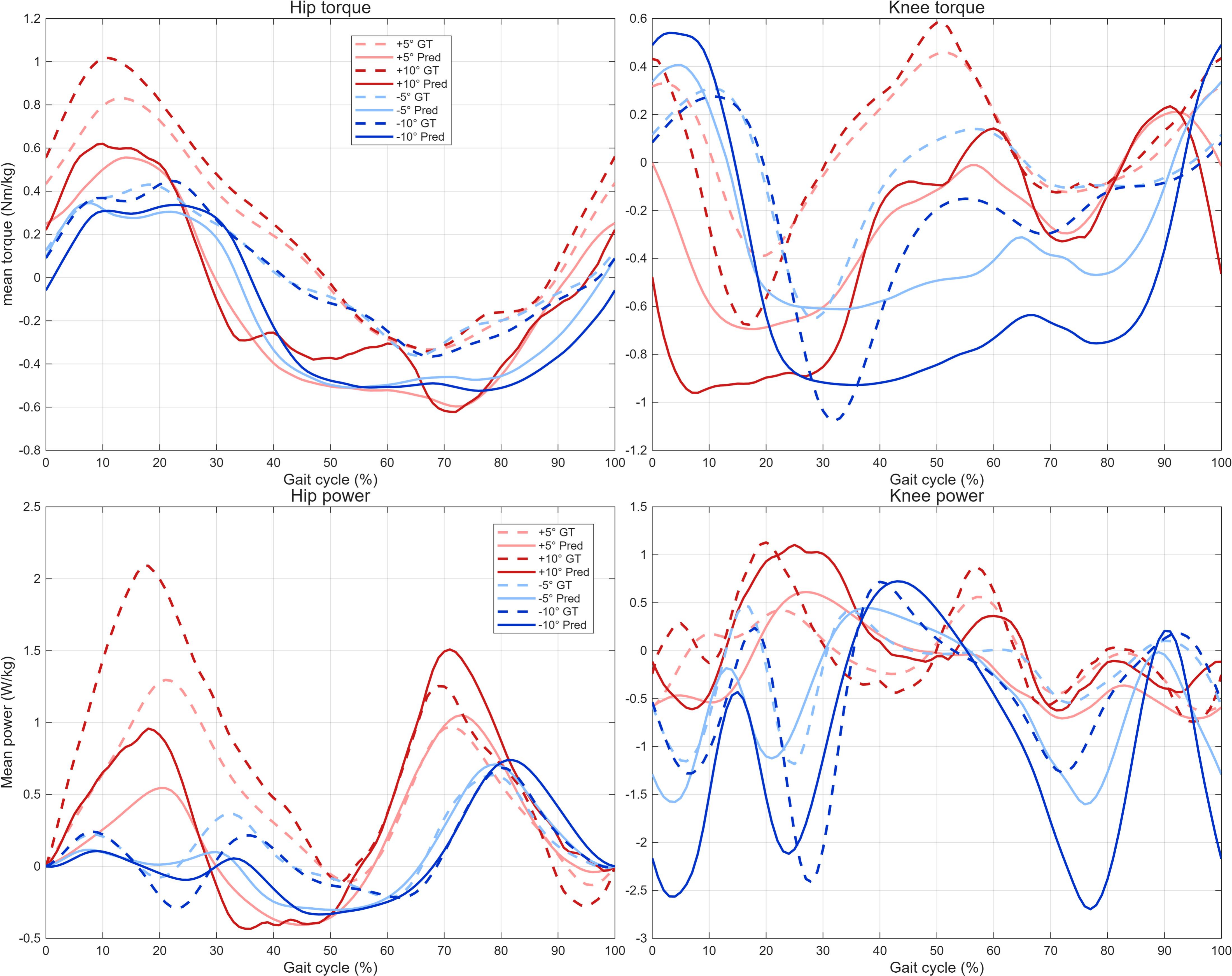}
  \caption{Inference hip and knee assistance torque and power generated from the level-ground walking model vs. GT biological moment and power under ramp walking. This is a mismatched model-task condition to examine the generalization capability of the level-ground walking model.}
  \label{fig:incline_walk_mismatch}
  \vspace{-0.2in}
\end{figure}

Fig.~\ref{fig:incline_walk_mismatch} compares the inferred assistance torques and power obtained from the level-ground model during ramp walking with the corresponding GT moments, which is a mismatached model and task condition. 
Overall, the inferred curves preserved a waveform trend similar to the GT but exhibited systematic underestimation in both magnitude and amplitude. 
For the hip torque, the incline trials showed clear underestimation of the early positive peak relative to GT. During the transition from positive peak to negative, a gradual decline was observed within the range of 35-70\%. 
In contrast, under other conditions, the inferred hip moment profiles closely resembled those observed under the level-ground walking condition, although both positive and negative peaks were reduced in magnitude. 
A similar pattern was observed for the knee joint. 
The inferred knee torque curves showed a strong and persistent negative bias, indicating larger mismatch. The -10° trial demonstrated a relatively long negative torque region and a late recovery toward the end of the cycle. At +10°, the knee moment showed the early-stance negative plateau (5–30\%).
The inferred hip power exhibited a negative power plateau between 35-50\% of the gait cycle for the +10° condition. Under the -10° condition, the knee power under -10° displayed multiple large negative absorption peaks, indicating that decline conditions amplified knee absorption-dominant errors when a level-ground policy was applied.

Table~\ref{tab:table1} presents the cross-correlation results between the predicted assistance torques and GT biological moments and joint power across all walking conditions. Overall, hip joint torque exhibited the highest consistency during level-ground walking, exceeding 0.9 under most conditions except at the lowest speed. In contrast, knee joint torque showed weaker correlation at low walking speeds but improved under ramp conditions, with cross-correlation values generally around 0.7. For hip joint power, the cross-correlation decreased from 0.90 to 0.61 as walking speed increased during level-ground walking. Knee joint power demonstrated relatively stable performance across level-ground conditions, while showing lower consistency during decline walking compared with incline walking.

\begin{figure}[htbp]
  \centering
  \includegraphics[width=1\linewidth]{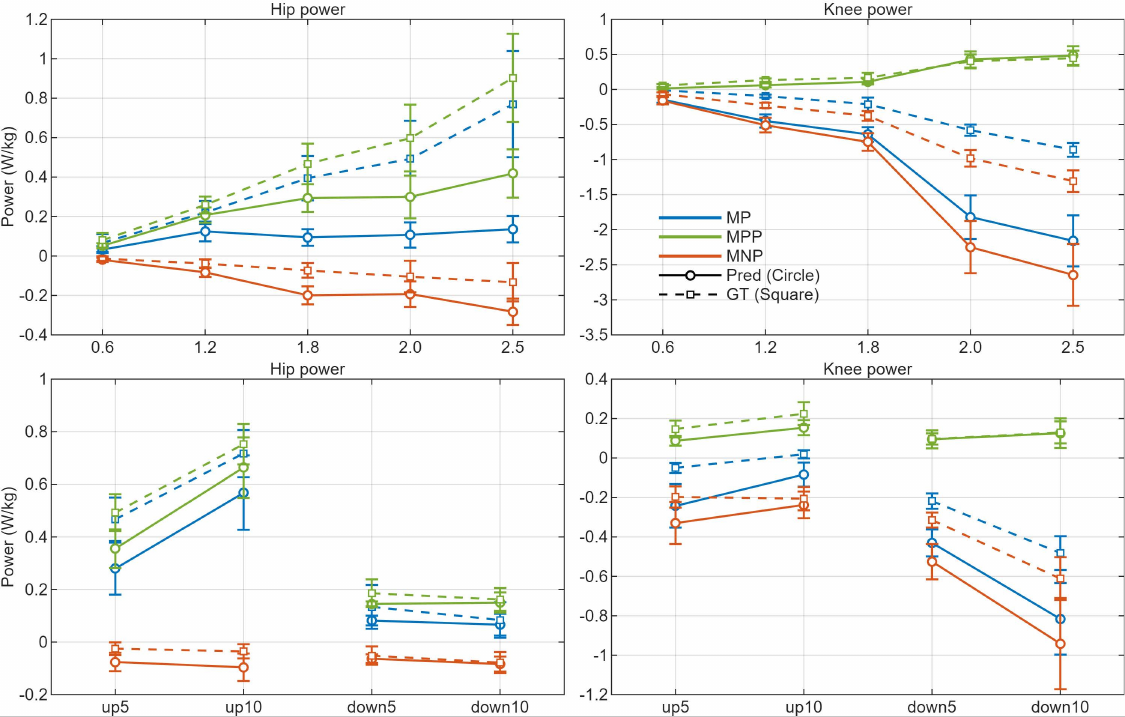}
  \caption{The means of inferred joint power of hip and knee, including MP, MPP and MNP. Top: walking under different speeds. Bottom: ramp walking at different incline/decline angles.}
  \label{fig:statistics}
  \vspace{-0.2in}
\end{figure}

Fig.~\ref{fig:statistics} presents the mean and standard deviation of the joint power under different walking speeds and ramp conditions.
For hip joint power, both GT and predicted MPP increased with walking speed. However, the predicted MPP consistently remained below GT, with larger discrepancies observed at 2\,m/s and 2.5\,m/s. In addition, compared with GT, the predicted MNP became progressively more negative as speed increases, resulting in a smaller overall increase in the predicted MP. 
For the knee joint, the MPP also increased with speed. The primary deviation arised in the MNP, where the predicted values declined sharply at higher speeds (2\,m/s and 2.5\,m/s), leading to more negative predicted MP compared with GT. The error bars further indicated increased inter-subject variability at higher speeds, particularly in the knee joint power metrics.
During ramp walking, for the hip joint, both GT and predicted MPP were higher under incline conditions than decline conditions, with the largest discrepancy between prediction and GT observed at +10°. Additionally, during incline walking, the predicted MNP was more negative than GT, resulting in a lower predicted MP overall.
For the knee joint, compared with GT, the -10° condition yielded the lowest predicted MP and MNP. Across all incline conditions, ±10° exhibited the greatest inter-subject variability, as reflected by the wider error bars in both joints. Despite these amplitude differences, the overall trend of power variation from different velocities and incline or decline grades were consistent between the predicted and GT data.




\begin{table*}[t]
\centering
\caption{Cross-correlation results and environment-wise averages for joint torque and power}
\vspace{-12pt}
\label{tab:table1}
\setlength{\tabcolsep}{4pt}
\renewcommand{\arraystretch}{1.15}

\subfloat[Cross-correlation between Inference and GT across walking conditions.]{
\label{tab:table1a}
\resizebox{0.63\textwidth}{!}{%
\begin{tabular}{lccccccccc}
\toprule
\textbf{ } &
\textbf{0.6\,m/s} & \textbf{1.2\,m/s} & \textbf{1.8\,m/s} & \textbf{2.0\,m/s} & \textbf{2.5\,m/s} &
\bm{$+5^\circ$} & \bm{$+10^\circ$} & \bm{$-5^\circ$} & \bm{$-10^\circ$} \\
\midrule
Hip torque  & 0.78 & 0.92 & 0.94 & 0.92 & 0.91 & 0.97 & 0.96 & 0.98 & 0.97 \\
Knee torque & 0.27 & 0.53 & 0.54 & 0.61 & 0.61 & 0.71 & 0.66 & 0.76 & 0.71 \\
Hip power   & 0.90 & 0.89 & 0.83 & 0.67 & 0.61 & 0.88 & 0.86 & 0.75 & 0.69 \\
Knee power  & 0.68 & 0.61 & 0.56 & 0.63 & 0.60 & 0.76 & 0.62 & 0.54 & 0.48 \\
\bottomrule
\end{tabular}%
}}%
\hfill
\subfloat[Mean values across three environments.]{
\label{tab:table1b}
\resizebox{0.34\textwidth}{!}{%
\begin{tabular}{lccc}
\toprule
\textbf{ } & \textbf{Level-ground} & \textbf{Incline} & \textbf{Decline} \\
\midrule
Hip torque  & 0.89 & 0.97 & 0.98 \\
Knee torque & 0.51 & 0.68 & 0.74 \\
Hip power   & 0.78 & 0.87 & 0.72 \\
Knee power  & 0.62 & 0.69 & 0.51 \\
\bottomrule
\end{tabular}%
}}
\vspace{-0.2in}
\end{table*}

\begin{figure}[htbp]
  \centering
  \includegraphics[width=1\linewidth]{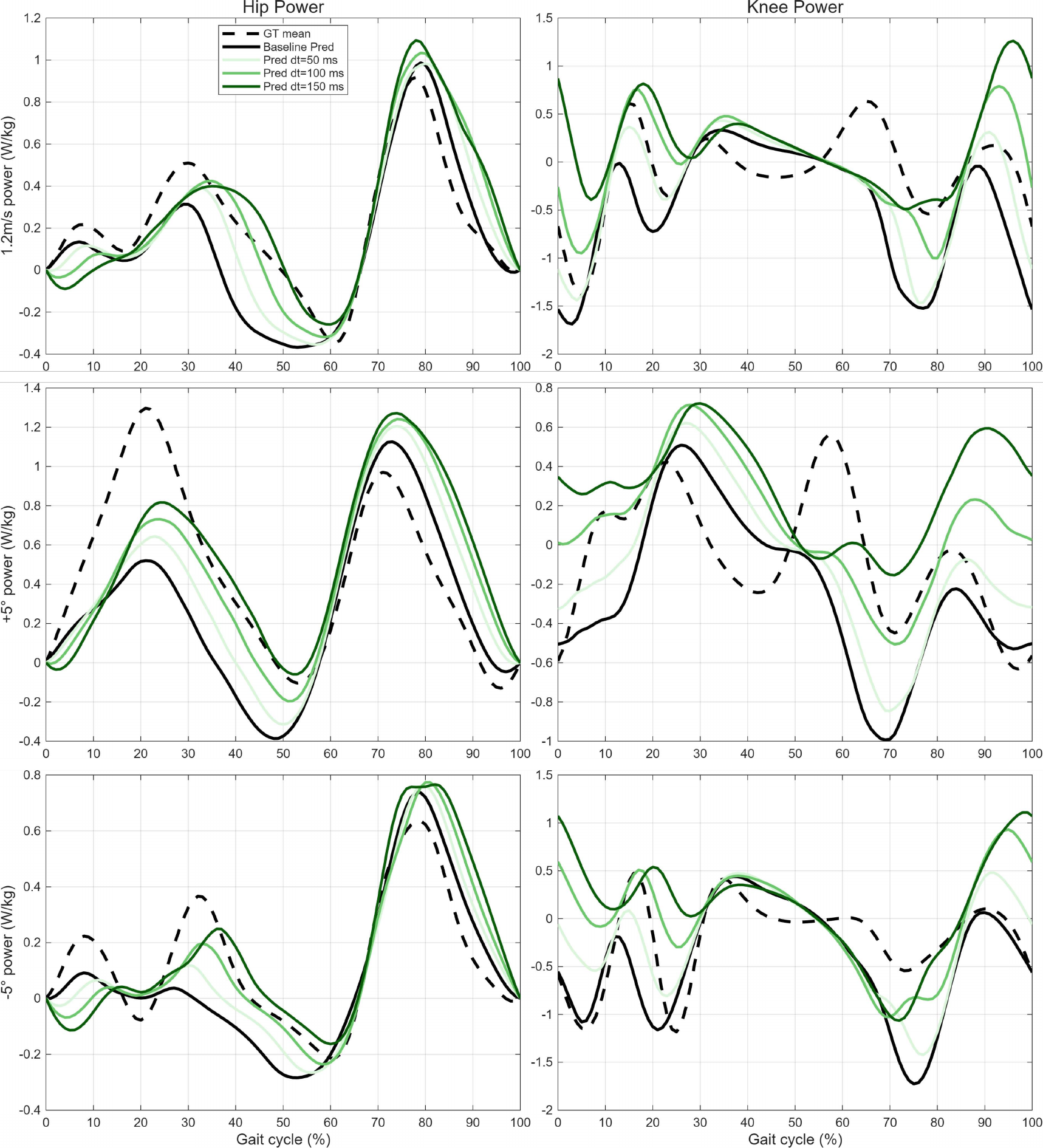}
  \caption{Inference hip and knee assistance power under different delay time vs. GT power under level-ground walking, inclined walking, and decline walking. Darker shades of green indicate longer delay time.}
  \label{fig:walk_power_delay}
  \vspace{-0.2in}
\end{figure}

Fig.~\ref{fig:walk_power_delay} presents, from top to bottom, the inferred joint power profiles obtained during level-ground walking at 1.2\,m/s, 5° incline walking, and 5° decline walking after introducing delays of 50\,ms, 100\,ms, and 150\,ms to the assistance torques (Pred dt). These results are compared with the zero-delay condition (Baseline Pred) and GT. 
Under level-ground walking at 1.2\,m/s, adding a 50–150\,ms delay on the assistance torques systematically reshaped the predicted joint power. Although the power profiles largely preserved the overall waveform structure, the most significant effect of introducing delay was not only elevating the positive power level but also reducing the magnitude of the power absorption peak, compared with the baseline results.
Compared with the zero-delay baseline, positive power peaks were amplified, whereas negative power peaks are reduced. 
For the hip joint, when the delay was set to 100 or 150\,ms, the variation during mid stance aligns more closely with GT, showing improvement compared with the zero delay condition. For the knee joint, the impact of delay is more evident. The delay not only increased the positive power level but also reduced the negative peaks compared to the baseline. Although larger delays brought the inferred curves closer to the GT profile during mid-gait, they produced an excessive upward trend in the later phase of the cycle.

For 5° incline walking, the introduction of torque delay consistently shifts the inferred joint power toward a more positive power–generating status. The most visible outcome of delay is an global amplification of predicted positive power and reduces the excessive negative peaks observed under the zero-delay baseline condition. The delayed curves that more closely matched the GT mean in both magnitude and overall energetic emphasis. This improvement was especially apparent at the hip, where delay acts primarily as an amplitude correction that increased the prominence of the main positive episode and mitigated excessive negative bias. At the knee, delay similarly reduces deep absorption artifacts and strengthened positive output. Although the delayed profiles appear smoother than the GT curves, the introduction of delay does not fully restore the knee joint power pattern.

During 5° decline walking, introducing torque delays again elevated the inferred power profiles, but the resulting changes were more mixed given that decline gait is characterized by substantial energy absorption rather than net generation. The most notable delay effect was a progressive suppression of negative power absorption paired with an increased positive excursion, which reduced the baseline’s deep absorption artifacts but simultaneously shifted the predicted energetic balance toward power generation. This trade-off is comparatively modest at the hip, where the delay brings the negative peak closer to the GT profile while augmenting the primary positive power phase. In contrast, the delay effect is more obvious at the knee. Larger delays (100\,ms and 150\,ms) markedly reduce the absorption-dominated characteristics and induce a late-stance positive power rise that deviates from the GT mean.

\section{DISCUSSION}
Fig.~\ref{fig:normal_walk_torque_power} indicates that, under level-ground walking conditions, the inferred hip and knee assistance torques preserve the expected speed-dependent modulation, with higher speeds producing systematically larger torque magnitudes and a consistent ordering across conditions. Secondly, the model effectively captures the amplitude scaling and the phase of extreme values in the joint moment profiles, indicating that it has learned the primary biomechanical mechanisms and the timing of key gait events. The hip exhibits more stable waveform similarity across speeds than the knee, suggesting that hip kinetics may serve as a more immediately reliable feedback signal for real-time assistance, while knee dynamics represent a clear target for subsequent refinement. Finally, the power profiles retain the correct directionality of energy exchange and show speed-amplified power events, supporting the utility of the inferred outputs for condition-aware assistance scaling and energy-based analysis.


The ramp walking results in Fig.~\ref{fig:incline_walk_torque_power} indicate that, both hip and knee outputs preserve the expected grade-dependent separation, consistently distinguishing incline from decline conditions and maintaining a coherent ordering across ±5° and ±10°. The level-ground walking results exhibit similar behavior, with dominant extrema in both assistance torque and joint power occurring within comparable phase windows to the GT, indicating that the model preserves the temporal sequence of key biomechanical events. Notably, the hip moment profiles remain relatively stable across different incline conditions, suggesting that hip dynamics provide a more robust foundation for real-time assistance, whereas knee dynamics still warrant further refinement. In the power profiles, the hip demonstrates more pronounced positive power generation during incline walking, while the knee exhibits stronger negative power absorption during decline walking. These trends are consistent with established energetic characteristics of ramp gait at the macroscopic level.


Fig.~\ref{fig:incline_walk_mismatch} further illustrates the distribution-shift effect when applying mismatched model and task input data. 
All inferred curves preserve the incline-dependent ordering and retain the dominant event structure within their respective phase windows. This indicates that even under model mismatch, the level-ground model does not completely fail, as both hip and knee profiles still differentiate the overall characteristics of incline and decline walking. These findings suggest that the learned kinematics-to-dynamics mapping possesses a certain degree of transferability. Furthermore, the hip torque and power profiles maintain better morphological consistency compared with those shown in Fig.~\ref{fig:incline_walk_torque_power}. Therefore, we hypothesize that in ramp scenarios, if a functional closed-loop assistance strategy must be implemented initially, hip dynamics may serve as the preferred reference. Because the errors of the mismatched model are substantially amplified at ±10°, incline should be treated as a critical condition variable during model training, underscoring the necessity of condition-specific modeling for particular environmental contexts.

As summarized in Table~\ref{tab:table1}, the prediction of hip assistance torques exhibits consistently high agreement with biological joint moments across all walking conditions. Regardless of variations in speed or incline, the hip moment correlation coefficients remain at a high level and become even higher when aggregated under ramp walking conditions. This indicates that models trained for different tasks can reliably capture the hip-dominant dynamics of gait. Moreover, the consistency of joint assistance prediction is reduced at the lowest (0.6\,m/s) and highest (2.5\,m/s) walking speeds, which is likely attributable to the domain randomization strategy employed during training. Specifically, the randomized speed range was set between 0.65\,m/s and 2.0\,m/s, following a normal distribution. Consequently, the model was exposed more frequently to intermediate speeds, such as 1.2\,m/s and 1.8 \,m/s, whereas samples at the extreme ends of the spectrum were underrepresented. Notably, the maintained performance at 0.6\,m/s and 2.5\,m/s further indicates a certain degree of generalization beyond the training speed range.
At the joint power level, the correlations are generally lower than those of joint torques, but they follow an interpretable pattern. For instance, hip power achieves the highest correlation under incline walking, suggesting that the model more effectively reconstructs power profiles in scenarios with elevated energetic demands.
In contrast, the knee joint represents the primary limitation of the current framework, with mismatches more evident during downhill walking. Overall, these statistical findings demonstrate that the proposed approach achieves reliable hip joint kinetic prediction and maintains robustness across varying locomotion environments.

In Fig.~\ref{fig:walk_power_delay}, this inference test shows how torque delays affect power estimates. For both joints, applying a small temporal shift to the inferred torques produces systematic, repeatable changes in the power profiles. At the hip, the effect of delay is more consistent. Increasing delay raises the late-cycle positive peak and reduces the depth of the mid-cycle negative peak, while maintaining the overall waveform structure. The knee exhibits stronger sensitivity, but delay consistently reduces the most extreme absorption (negative power) event and increases the positive peak shifting, demonstrating that time alignment can reshape the knee power pattern to resemble the GT distribution better.

Similar benefits of temporal alignment persist on ramps, supporting the idea that timing compensation can transfer across terrains. On the 5° incline, delay strengthens the positive-power episodes and reduces excessive negative excursions relative to the zero-delay prediction, improving the balance between generation and absorption events that characterizes incline walking gait. On the 5° decline, delay again moderates the depth and duration of the knee absorption trough and enhances the late-cycle recovery, which is especially valuable because decline walking amplifies braking-dominated energy absorption. Across both ramp conditions, the delay effects are systematic but not identical. The results consistently show that a modest torque delay can bias the reconstructed joint power toward increased positive power and reduced absorption.

This study has two main limitations. First, the knee joint demonstrates comparatively lower inferential consistency, which may stem from structural simplifications in the simulation model. Specifically, the knee is represented as a simple hinge joint rather than a translation–rotation coupled mechanism, thereby neglecting secondary motions and complex joint interactions that influence physiological behavior~\cite{scott1999interactive}. More broadly, accurately reproducing knee mechanics, including both kinematic and kinetic variables, remains a persistent challenge in predictive musculoskeletal simulations due to the anatomical complexity of the tibiofemoral and patellofemoral joints, sensitivity to modeling assumptions, and control formulation choices~\cite{de2021perspective,afschrift2025benchmarking}. 
Secondly, the estimated power is sensitive to timing shift between inferred torque and joint angular velocity. The delay experiments demonstrate that modest timing shifts systematically reshape key power events and that the delay yielding better agreement varies across terrains, implying that the observed difference between predicted and GT power are not solely attributable to torque magnitude prediction but also to timing mismatch. In future work, we will consider fine tuning the inference models with the delay parameter input to improve the accuracy of inferred assistance torques and timing shift under more complex conditions.


\section{CONCLUSIONS}
Our results demonstrate that a physics-based RL framework trained entirely in simulation can produce exoskeleton control policies that generate assistance torque resembling and reducing human biological joint moments. The learned policy produces stable and biomechanically interpretable assistance patterns for both hip and knee joints across diverse walking conditions, validated using an open-source gait dataset. Hip assistance, in particular, shows stronger generalization and higher similarity to biological joint moments, with torque and power profiles that better preserve key temporal structure and amplitude trends compared to those of the knee. These findings suggest that simulation-learned hip support is a robust component for unified assistive control.
On the other hand, the knee assistance is more sensitive to distribution shifts and temporal misalignment, highlighting the need for broader training coverage and improved modeling strategies. 
We further observe that applying a level-ground-trained controller to ramp walking introduces systematic, phase-dependent distortions, underscoring that condition-mismatched models should not be treated as default solutions. 
Delay analysis further shows that modest timing shifts can reshape power events and improve phase-specific alignment, indicating that controlled torque delays may serve as a practical mechanism to bias assistance toward more effective positive power injection. Future work will focus on deploying the trained controllers on physical exoskeleton platforms and collecting data from a larger cohort of human subjects under real operating conditions to enable comprehensive validation, refinement, and assessment of sim-to-real transfer performance.

\bibliographystyle{IEEEtran}
\bibliography{PaperReference}

@article{lim2019prediction,
  title={Prediction of lower limb kinetics and kinematics during walking by a single IMU on the lower back using machine learning},
  author={Lim, Hyerim and Kim, Bumjoon and Park, Sukyung},
  journal={Sensors},
  volume={20},
  number={1},
  pages={130},
  year={2019},
  publisher={MDPI}
}

@article{leem2026exo,
  title={Exo-Plore: Exploring Exoskeleton Control Space through Human-aligned Simulation},
  author={Leem, Geonho and Lee, Jaedong and Lee, Jehee and Song, Seungmoon and Won, Jungdam},
  journal={arXiv preprint arXiv:2601.22550},
  year={2026}
}

@article{schulman2017proximal,
  title={Proximal policy optimization algorithms},
  author={Schulman, John and Wolski, Filip and Dhariwal, Prafulla and Radford, Alec and Klimov, Oleg},
  journal={arXiv preprint arXiv:1707.06347},
  year={2017}
}

@article{lee2019scalable,
  title={Scalable muscle-actuated human simulation and control},
  author={Lee, Seunghwan and Park, Moonseok and Lee, Kyoungmin and Lee, Jehee},
  journal={ACM Transactions On Graphics (TOG)},
  volume={38},
  number={4},
  pages={1--13},
  year={2019},
  publisher={ACM New York, NY, USA}
}

@article{belal2024deep,
  title={Deep learning approaches for enhanced lower-limb exoskeleton control: A review},
  author={Belal, Mohammad and Alsheikh, Nael and Aljarah, Ahmad and Hussain, Irfan},
  journal={Ieee Access},
  volume={12},
  pages={143883--143907},
  year={2024},
  publisher={IEEE}
}

@article{baud2021review,
  title={Review of control strategies for lower-limb exoskeletons to assist gait},
  author={Baud, Romain and Manzoori, Ali Reza and Ijspeert, Auke and Bouri, Mohamed},
  journal={Journal of neuroengineering and rehabilitation},
  volume={18},
  number={1},
  pages={119},
  year={2021},
  publisher={Springer}
}

@article{scott1999interactive,
  title={An Interactive Graphics-Based Model of the Lower Extremity to Study Orthopaedic Surgical},
  author={SCOTT, L and LOAN, J PETER and MELISSA, G and I-IOY, FELIX E ZAJAC and ERIC, L and JOSEPH, M ROSEN},
  journal={IEEE TRANSACTIONS ON BIOMEDICAL ENGINEERING},
  volume={3},
  number={3},
  pages={75},
  year={1999}
}

@article{zhang2025joint,
  title={Joint moment estimation for hip exoskeleton control: A generalized moment feature generation method},
  author={Zhang, Yuanwen and Xiong, Jingfeng and Xian, Haolan and Chen, Chuheng and Chen, Xinxing and Liang, Haipeng and Fu, Chenglong and Leng, Yuquan},
  journal={Biomimetic Intelligence and Robotics},
  pages={100246},
  year={2025},
  publisher={Elsevier}
}

@article{molinaro2024estimating,
  title={Estimating human joint moments unifies exoskeleton control, reducing user effort},
  author={Molinaro, Dean D and Kang, Inseung and Young, Aaron J},
  journal={Science robotics},
  volume={9},
  number={88},
  pages={eadi8852},
  year={2024},
  publisher={American Association for the Advancement of Science}
}

@inproceedings{molinaro2020biological,
  title={Biological hip torque estimation using a robotic hip exoskeleton},
  author={Molinaro, Dean D and Kang, Inseung and Camargo, Jonathan and Young, Aaron J},
  booktitle={2020 8th IEEE RAS/EMBS International Conference for Biomedical Robotics and Biomechatronics (BioRob)},
  pages={791--796},
  year={2020},
  organization={IEEE}
}

@article{su2023simulating,
  title={Simulating human walking: a model-based reinforcement learning approach with musculoskeletal modeling},
  author={Su, Binbin and Gutierrez-Farewik, Elena M},
  journal={Frontiers in neurorobotics},
  volume={17},
  pages={1244417},
  year={2023},
  publisher={Frontiers Media SA}
}

@article{altai2023performance,
  title={Performance of multiple neural networks in predicting lower limb joint moments using wearable sensors},
  author={Altai, Zainab and Boukhennoufa, Issam and Zhai, Xiaojun and Phillips, Andrew and Moran, Jason and Liew, Bernard XW},
  journal={Frontiers in bioengineering and biotechnology},
  volume={11},
  pages={1215770},
  year={2023},
  publisher={Frontiers Media SA}
}

@article{liang2023deep,
  title={Deep-learning model for the prediction of lower-limb joint moments using single inertial measurement unit during different locomotive activities},
  author={Liang, Wenqi and Wang, Fanjie and Fan, Ao and Zhao, Wenrui and Yao, Wei and Yang, Pengfei},
  journal={Biomedical Signal Processing and Control},
  volume={86},
  pages={105372},
  year={2023},
  publisher={Elsevier}
}

@article{de2021deep,
  title={Deep reinforcement learning for physics-based musculoskeletal simulations of healthy subjects and transfemoral prostheses’ users during normal walking},
  author={De Vree, Leanne and Carloni, Raffaella},
  journal={IEEE Transactions on Neural Systems and Rehabilitation Engineering},
  volume={29},
  pages={607--618},
  year={2021},
  publisher={IEEE}
}

@article{huang2025lower,
  title={A lower limb exoskeleton adaptive control method based on model-free reinforcement learning and improved dynamic movement primitives},
  author={Huang, Liping and Zheng, Jianbin and Gao, Yifan and Song, Qiuzhi and Liu, Yali},
  journal={Journal of Intelligent \& Robotic Systems},
  volume={111},
  number={1},
  pages={24},
  year={2025},
  publisher={Springer}
}

@article{zhang2024integrating,
  title={Integrating musculoskeletal simulation and machine learning: a hybrid approach for personalized ankle-foot exoskeleton assistance strategies},
  author={Zhang, Xianyu and Li, Shihao and Ying, Zhenzhi and Shu, Liming and Sugita, Naohiko},
  journal={Frontiers in Bioengineering and Biotechnology},
  volume={12},
  pages={1442606},
  year={2024},
  publisher={Frontiers Media SA}
}

@article{rose2022model,
  title={A model-free deep reinforcement learning approach for control of exoskeleton gait patterns},
  author={Rose, Lowell and Bazzocchi, Michael CF and Nejat, Goldie},
  journal={Robotica},
  volume={40},
  number={7},
  pages={2189--2214},
  year={2022},
  publisher={Cambridge University Press}
}

@article{luo2024experiment,
  title={Experiment-free exoskeleton assistance via learning in simulation},
  author={Luo, Shuzhen and Jiang, Menghan and Zhang, Sainan and Zhu, Junxi and Yu, Shuangyue and Dominguez Silva, Israel and Wang, Tian and Rouse, Elliott and Zhou, Bolei and Yuk, Hyunwoo and others},
  journal={Nature},
  volume={630},
  number={8016},
  pages={353--359},
  year={2024},
  publisher={Nature Publishing Group UK London}
}

@article{luo2021reinforcement,
  title={Reinforcement learning and control of a lower extremity exoskeleton for squat assistance},
  author={Luo, Shuzhen and Androwis, Ghaith and Adamovich, Sergei and Su, Hao and Nunez, Erick and Zhou, Xianlian},
  journal={Frontiers in Robotics and AI},
  volume={8},
  pages={702845},
  year={2021},
  publisher={Frontiers Media SA}
}

@article{luo2023robust,
  title={Robust walking control of a lower limb rehabilitation exoskeleton coupled with a musculoskeletal model via deep reinforcement learning},
  author={Luo, Shuzhen and Androwis, Ghaith and Adamovich, Sergei and Nunez, Erick and Su, Hao and Zhou, Xianlian},
  journal={Journal of neuroengineering and rehabilitation},
  volume={20},
  number={1},
  pages={34},
  year={2023},
  publisher={Springer}
}

@inproceedings{park2022generative,
  title={Generative gaitnet},
  author={Park, Jungnam and Min, Sehee and Chang, Phil Sik and Lee, Jaedong and Park, Moon Seok and Lee, Jehee},
  booktitle={ACM SIGGRAPH 2022 Conference Proceedings},
  pages={1--9},
  year={2022}
}

@article{molinaro2024task,
  title={Task-agnostic exoskeleton control via biological joint moment estimation},
  author={Molinaro, Dean D and Scherpereel, Keaton L and Schonhaut, Ethan B and Evangelopoulos, Georgios and Shepherd, Max K and Young, Aaron J},
  journal={Nature},
  volume={635},
  number={8038},
  pages={337--344},
  year={2024},
  publisher={Nature Publishing Group UK London}
}

@article{afschrift2025benchmarking,
  title={Benchmarking the predictive capability of human gait simulations},
  author={Afschrift, Maarten and Kistemaker, Dinant and Bobbert, Maarten and De Groote, Friedl},
  journal={PLOS Computational Biology},
  volume={21},
  number={11},
  pages={e1012713},
  year={2025},
  publisher={Public Library of Science San Francisco, CA USA}
}

@article{de2021perspective,
  title={Perspective on musculoskeletal modelling and predictive simulations of human movement to assess the neuromechanics of gait},
  author={De Groote, Friedl and Falisse, Antoine},
  journal={Proceedings of the Royal Society B: Biological Sciences},
  volume={288},
  number={1946},
  year={2021},
  publisher={The Royal Society}
}

@article{mccabe2023developing,
  title={Developing a method for quantifying hip joint angles and moments during walking using neural networks and wearables},
  author={McCabe, Megan V and Van Citters, Douglas W and Chapman, Ryan M},
  journal={Computer Methods in Biomechanics and Biomedical Engineering},
  volume={26},
  number={1},
  pages={1--11},
  year={2023},
  publisher={Taylor \& Francis}
}

\end{document}